%% file: example_paper.tex
\documentclass[nohyperref]{article}

\usepackage{siyu}
\usepackage{microtype}
\usepackage{graphicx}
\usepackage{subfigure}
\usepackage{booktabs} % for professional tables

\usepackage{hyperref}

\usepackage[accepted]{icml2022}

\usepackage{amsmath}
\usepackage{amssymb}
\usepackage{mathtools}
\usepackage{amsthm}

\usepackage[capitalize,noabbrev]{cleveref}

\theoremstyle{plain}

\theoremstyle{definition}

\theoremstyle{remark}

\usepackage{ifthen}
\newif\ifdebug
\debugtrue
\ifdebug
\usepackage[textsize=tiny,color=lightgray,textwidth=2cm]{todonotes}
\else
\usepackage[disable]{todonotes}
\fi

\definecolor{babypink}{rgb}{0.96, 0.76, 0.76}

\icmltitlerunning{Fast Lossless Neural Compression with Integer-Only Discrete Flows}

\begin{document}

\twocolumn[
\icmltitle{Fast Lossless Neural Compression with Integer-Only Discrete Flows}

% It is OKAY to include author information, even for blind
% submissions: the style file will automatically remove it for you
% unless you've provided the [accepted] option to the icml2022
% package.

% List of affiliations: The first argument should be a (short)
% identifier you will use later to specify author affiliations
% Academic affiliations should list Department, University, City, Region, Country
% Industry affiliations should list Company, City, Region, Country

% You can specify symbols, otherwise they are numbered in order.
% Ideally, you should not use this facility. Affiliations will be numbered
% in order of appearance and this is the preferred way.
\icmlsetsymbol{equal}{*}

\begin{icmlauthorlist}
\icmlauthor{Siyu Wang}{thu}
\icmlauthor{Jianfei Chen}{thu}
\icmlauthor{Chongxuan Li}{ruc}
\icmlauthor{Jun Zhu}{thu}
\icmlauthor{Bo Zhang}{thu}
%\icmlauthor{}{sch}
%\icmlauthor{}{sch}
\end{icmlauthorlist}

\icmlaffiliation{thu}{Dept. of Comp. Sci. \& Tech., BNRist Center, Tsinghua-Bosch Joint ML Center, Tsinghua University; Peng Cheng Laboratory} % to modify
\icmlaffiliation{ruc}{Gaoling School of AI, Renmin University of China; Beijing Key Lab of Big Data Management \& Analysis Methods, Beijing, China}

\icmlcorrespondingauthor{Jianfei Chen}{jianfeic@tsinghua.edu.cn}
\icmlcorrespondingauthor{Jun Zhu}{dcszj@tsinghua.edu.cn}

% You may provide any keywords that you
% find helpful for describing your paper; these are used to populate
% the "keywords" metadata in the PDF but will not be shown in the document
\icmlkeywords{Machine Learning, ICML}

\vskip 0.3in
]

\printAffiliationsAndNotice{}  % leave blank if no need to mention equal contribution
% \printAffiliationsAndNotice{\icmlEqualContribution} % otherwise use the standard text.

\begin{abstract}
By applying entropy codecs with learned data distributions, neural compressors have significantly outperformed traditional codecs in terms of compression ratio. 
However, the high inference latency of neural networks hinders the deployment of neural compressors in practical applications. 
In this work, we propose Integer-only Discrete Flows (\modelm), an efficient neural compressor with integer-only arithmetic. Our work is built upon integer discrete flows, which consists of invertible transformations between discrete random variables. We propose efficient invertible transformations with integer-only arithmetic based on 8-bit quantization. Our invertible transformation is equipped with learnable binary gates to remove redundant filters during inference. 
We deploy \model with TensorRT on GPUs, achieving $10\times$ inference speedup compared to the fastest existing neural compressors, while retaining the high compression rates on ImageNet32 and ImageNet64. 
   
\end{abstract}

\section{Introduction}

As a growing amount of data is produced every day, efficient lossless compression is of significance in storing and transmitting them. Shannon's source coding theorem \cite{shannon1948mathematical} states that the average code length needed to encode data is lower bounded by entropy of its distribution:
\begin{equation}
    \EXP_{\xx\sim p_\cD}[|c(\xx)|] \geq \EXP_{\xx \sim p_\cD}[-\log p_\cD(\xx)],
\end{equation}
where $|c(\xx)|$ is the code length and $p_\cD$ is the data distribution.  Based on the insight that the optimal code length for a single symbol $\xx$ is $-\log p_\cD(\xx)$, many efficient entropy codecs \cite{huffman1952method, duda2009asymmetric, duda2013asymmetric} have been developed, and they have achieved nearly optimal code length given known data distribution. However, the data distribution is generally unknown in practice. 

\begin{figure}[ht]
\vskip 0.2in
\begin{center}
\centerline{\includegraphics[width=\columnwidth]{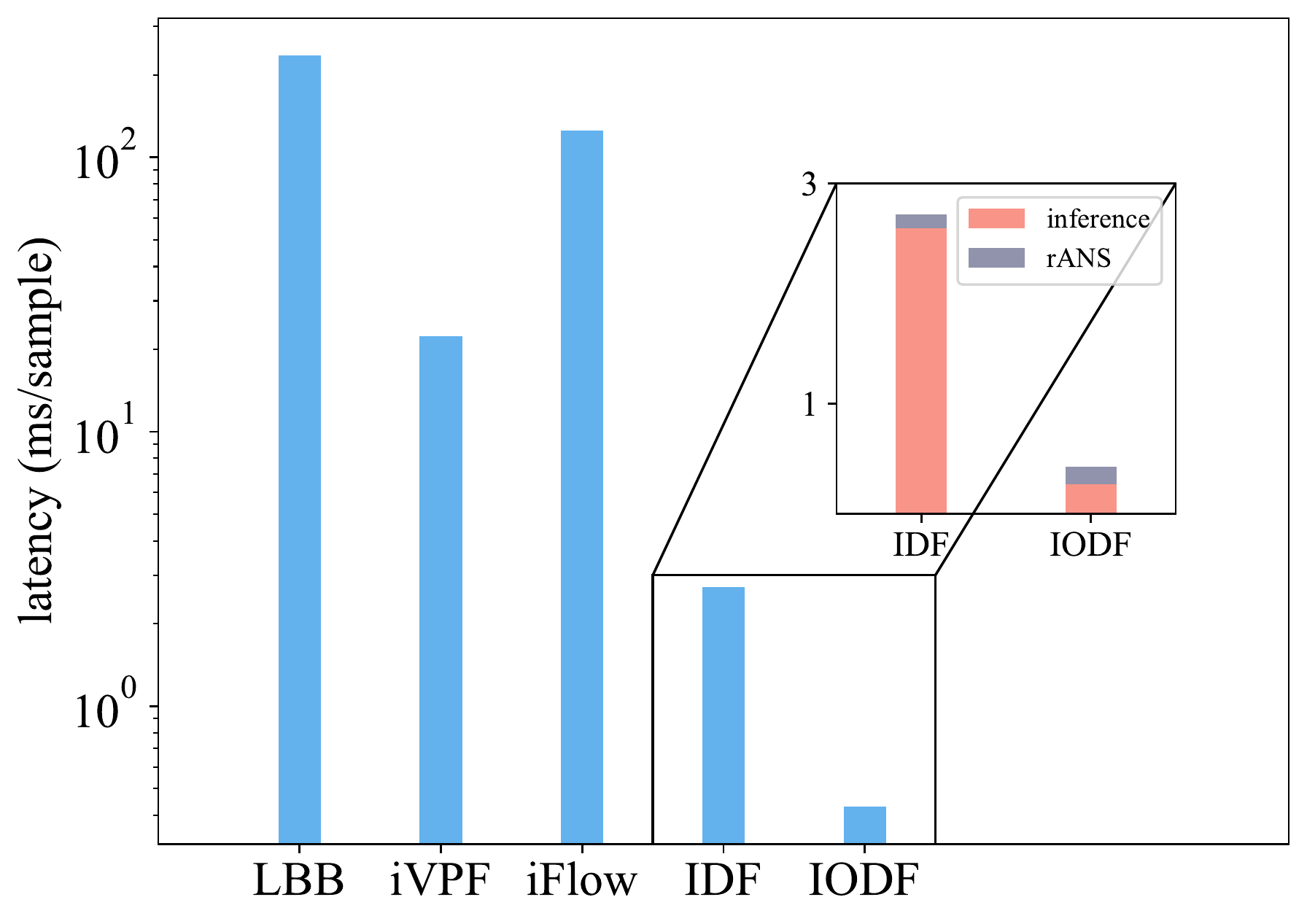}}
\caption{Encoding latency of different neural compressors measured by milliseconds per sample. Statistics of LBB, iVPF, and iFlow are directly picked from \cite{zhang2021ivpf, zhang2021iflow}. The inner figure shows inference and entropy coding latency of IDF and \model separately. For IDF,  inference is the time bottleneck in the encoding process. \model significantly improves the inference speed, making it comparable to entropy coding.}
\label{fig:dgm_comp}
\end{center}
\vskip -0.2in
\end{figure}

In the machine learning community, various likelihood-based generative models have been developed, including normalizing flows (\citet{dinh2016density, ho2019flow++, chen2020vflow, lu2021implicit}, NFs), variational auto-encoders (\citet{Kingma2014AutoEncodingVB, rezende2014stochastic}, VAEs), auto-regressive models (\citet{4587817, larochelle2011neural, salimans2017pixelcnn++}, ARMs), and diffusion models (\citet{sohl2015deep, ho2020denoising, song2020denoising}, DPMs). 
These deep generative models (DGMs) are powerful density estimators. %Equipped with learned data distributions,
With a model distribution close enough to the data distribution, 
many efficient lossless \emph{neural compressors} have been explored \cite{hoogeboom2019integer, van2020idf++, ho2019compression, townsend2019practical,  kingma2021variational}. Neural compressors have achieved superior compression ratios on standard datasets than traditional lossless compression methods, such as JPEG2000 \cite{joint2000jpeg2000}, PNG \cite{boutell1997png} and ELIF \cite{sneyers2016flif}. 

Despite the high compression rates, existing neural compressors still suffer from a low coding bandwidth, which has hindered practical applications. 
% due to their low coding bandwidth. %compared to traditional lossless compression methods. 
Figure~\ref{fig:dgm_comp} shows the encoding latency of several NFs-based neural compressors on the ImageNet32 dataset, of which Integer Discrete Flows (IDF) \cite{hoogeboom2019integer} is the fastest. 
However, the coding bandwidth of IDF is still more than an order of magnitude slower than traditional image codecs such as JPEG2000.

Technically, by viewing both image data and latent variables in a discrete integer space, IDF designs a bijective mapping between them for exact likelihood inference. Then it utilizes the rANS \cite{duda2009asymmetric, duda2013asymmetric} coding algorithm to encode images into bits stream. Inference and entropy encoding are two separate steps of compression with IDF. Figure~\ref{fig:dgm_comp} illustrates that inference is over ten times slower than entropy coding for IDF. Thus, we seek to improve the coding bandwidth by accelerating the inference of flow-based models.

In this work, we present integer-only discrete flow (\modelm), an efficient discrete flow for neural compression.
We leverage quantization 
methods~\citep{jacob2018quantization,esser2019learned} to accelerate the inference. 
Unlike IDF, which defines discrete bijections with expensive continuous neural networks, \model performs its basic operations with the efficient \emph{integer} arithmetic. Furthermore, \model is equipped with learnable binary gates to identify and prune out redundant computations during inference. Our experiments demonstrate that \model achieves up to 10$\times$ inference speedup compared to IDF. In summary, our contributions include:

\begin{itemize}    
\item We propose an efficient integer-only neural architecture for discrete flows. The architecture is carefully designed to be hardware-friendly to allow fast inference. 
We propose an algorithm to train such integer-only architectures, where \model achieves comparable density estimation performance with the full precision IDF.
\item We propose to prune IDF with learnable integer (more specifically, binary) gates. By removing redundant filters, we reduce FLOPs of IDF from $7.2$G to $3.2$G with only a tiny increase in bits per dimension (bpd).
\item We deploy \model with integer-only computational kernels on a Tesla T4 GPU using the TensorRT library~\cite{tensorrt}. We show that with integer arithmetic and pruning, \model can achieve up to $10\times$ speedup compared to IDF during inference. \model makes a step forward towards practical application of deep generative models in data compression. 
\end{itemize}

\section{Related Work}

{\textbf {Coding with DGMs}} 
Based on the \emph{change-of-variable formula}, NFs perform exact data distribution inference by designing bijective maps between data $\xx$ and latent representation $\zz$. Combining with entropy coding algorithms, NFs are applied to data lossless compression. 
Different from IDF \cite{hoogeboom2019integer, van2020idf++} which model $\xx$, $\zz$ both with discrete integers, \citet{ho2019compression} dicretize continuous variables and propose a \emph{local bits back} (LBB) scheme for compression with a general class of NFs which model $\xx$, $\zz$ as continuous. 
\citet{zhang2021ivpf, zhang2021iflow} aim to handle the problem that continuous NFs are numerically non-invertible in data compression, and they propose novel NFs with numerically invertible transformations.
Although these continuous flow-based models have achieved higher compression rates than IDF, their inference and coding procedures are generally more complex and slower. 
Currently, hierarchical VAEs attain theoretical code lengths comparable to NFs \cite{maaloe2019biva, ho2019flow++} and DPMs achieve the best code lengths \cite{kingma2021variational} (2.49 bits/dim on CIFAR10, 3.72 bits/dim on ImageNet32 and 3.40 bits/dim on ImageNet64).
However, the bandwidth of the corresponding compressors is much lower than IDF ($\sim$1MB/s) and far from practical demand, e.g., \citet{townsend2019hilloc} compresses at $\sim$ 0.02MB/s and \citet{kingma2021variational} requires expensive computation due to a large number of timesteps with a network forward in each timestep.
Moreover, continuous NFs, VAEs and DPMs based compression algorithms rely on the bits back coding scheme and thus suffer from non-negligible auxiliary bits when there are only a small amount of data to compress. 
Meanwhile, ARMs attain theoretical code lengths similar to DPMs, but decoding is extremely slow due to their serial sampling procedure. 
Overall, IDF is the most efficient neural compressor at the cost of a weak compression rate loss.

\textbf{Pruning and Quantization} 
Pruning and quantization are popular methods for reducing the memory and latency of deep neural networks (DNNs). Pruning strategies generally follow a procedure that first trains a network to convergence, scores the weight parameters (usually based on $l_1 / l_2$-norms), prunes out low-scored parameters, and fine-tunes the pruned networks \cite{lecun1990optimal, han2015learning, li2016pruning, lebedev2016fast, wen2016learning, he2017channel, frankle2018lottery}. Although various pruning methods have been developed for classification models, there is little attention on pruning NFs.
Quantization represents parameters and activations in low precision format instead of 32-bit floating-point numbers~\cite{courbariaux2015binaryconnect, courbariaux2016binarized, rastegari2016xnor,zhou2016dorefa, jacob2018quantization, choi2018pact, dong2019hawq, esser2019learned, chen2020statistical, van2020bayesian}. 
While researches in low-bit neural networks mainly focus on the quantization of discriminative models, there have been recent attempts to introduce quantization techniques into generative models. \citet{bird2020reducing} binarize the majority of weights and activations in deep hierarchical VAE and NFs while retaining a valid probabilistic model. However, binarizing weights and activations leads to significant performance degradation %(e.g., bits per dimension on ImageNet32x32 drops from 4.05 of 32-bit Flow++ to 4.30 of 1-bit Flow++).
Moreover, they use \emph{simulated quantization} and all operations are still performed in floating points, which can not improve inference speed in reality. 
\citet{balle2018integer} explores integer networks and quantization in generative models for lossy compression, aiming to address the problem that floating-point arithmetic are not deterministic across different platforms.

Briefly, current DGMs based neural compressors still suffer from high inference latency, among which IDF is the most time efficient. On the other hand, pruning and quantization techniques for NFs need further exploration. Our work aims to speed up inference of IDF with integer computations and pruning out redundant computations. 

\section{Background: Integer Discrete Flows}

\label{sec:idf}
Normalizing flows (NFs) provide a general framework for learning probability distributions over continuous and discrete random variables. 
In the discrete case, NFs consider $\xx$ to be a discrete random variable with unknown distribution $p_X(\xx)$. Then NFs construct an invertible transformation $f: \mathcal{X} \mapsto \mathcal{Z}$, mapping $\xx$ to latent representation $\zz = f(\xx)$ on which we impose a tractable density $p_Z(\zz)$. Then the density of $\xx$ can be obtained by the change-of-variables formula: 

\begin{equation*}
    p_X(\xx) = \sum_{z\in \{f\rev(z)=x\}} p_Z(\zz).
\end{equation*}
Consider that $f:\cX \mapsto \mathcal{Z}$ is invertible, the summation set contains only $\zz=f(\xx)$, so the probability mass of $\xx$ is given by 
\begin{equation*}
    p_X(\xx) = p_Z(\zz).
\end{equation*}

IDF \cite{hoogeboom2019integer} assumes that both $\xx$ and $\zz$ lie in the $d-$dimensional integer space so $\cX = \mathcal{Z} = \ints^d$, and the prior distribution $p_Z(\zz)$ is chosen as factorized discrete  logistic distribution in the form
\begin{equation*}
    p_Z(\zz|\bm{\mu},\bm{s}) = \prod_{i=1}^d \left(\sigma(\frac{z_i+\frac12-\mu_i}{s_i}) - \sigma(\frac{z_i-\frac12-\mu_i}{s_i})\right),
\end{equation*}
where $\sigma(\cdot)$ denotes the sigmoid function. To obtain an invertible function, IDF designs its basic building block as an additive coupling layer \cite{dinh2016density}:
\begin{equation}
    \label{eq:coupling}
    \left[\begin{array}{l}\zz_{a} \\ \zz_{b}\end{array}\right]=\left[\begin{array}{c}\xx_{a} \\ \xx_{b}+ \round{t_{\theta}\left(\xx_{a}\right)}\end{array}\right].
\end{equation}
Here $\xx$ is split into two parts $\xx_a \in \ints^m, \xx_b\in\ints^n$ with $m+n=d$; likewise for $\zz_a, \zz_b$. $t_\theta(\cdot)$ is a neural network  parameterized by $\theta$. The network $t_\theta(\cdot): \real^m \rightarrow \real^n$ defines a continuous mapping, which is projected to the discrete domain by the rounding operator $\round{\cdot}$. We defer the optimization of neural networks with the rounding operator to Sec.~\ref{sec:optim_quant}.

IDF defines a discrete invertible mapping with a continuous function $t_\theta(\cdot)$, but the network $t_\theta(\cdot)$ still operates in the continuous domain internally. This makes IDF slow since expensive float-point operations are performed within the network.

\section{Methodology}
To make neural compression algorithms more efficient, we propose integer-only discrete flows (\modelm). \model consists of a novel network architecture for $t_\theta(\cdot)$, where most of the computations are achieved by efficient integer operations. 
\model also prunes redundant convolution filters with learnable binary gates, which are again implemented with integer operations. We discuss how to train such integer-only networks. 
Finally, we propose hardware-friendly optimizations to maximize the efficiency of the integer arithmetic in \modelm, including a reconsideration of the backbone architecture and a carefully implemented shortcut path.

\subsection{Integer-Only Residual Block}
\label{sec:int_resb}
We first present the basic integer-only building block of IODF. The methodology is mostly based on existing works on neural network quantization~\citep{jacob2018quantization,esser2019learned}, but we present the details below in the normalizing flow context. 

In \modelm, each network $t_\theta(\cdot)$ is 
made of a sequence of $L$ \emph{integer-only residual blocks} $t_\theta(\xx) = t^{(1)}_\theta\circ \dots \circ t^{(L)}_\theta(\xx)$, where each block is defined as
\begin{equation}
    \label{eq:resb}
    t_\theta^{(l)}(\xx) = \relu(Q(\xx) + \conv(\relu(\conv(Q(\xx))))).
\end{equation}
Note that this ResNet-like architecture~\cite{he2016deep} differs from the DenseNet architecture~\cite{huang2017densely} used in IDF, which we will explain in Sec.~\ref{sec:arch}. 

In the integer-only residual block, all the tensors are represented with a hybrid numerical format, where a \emph{quantizer} $Q$ is used to convert floating-point tensors to the hybrid format. For a real-valued tensor $\rr$, the quantizer outputs
\begin{equation}
    \label{eq:q}
    \tilde{\rr} := Q(\rr) = s_\rr\hat{\rr} \approx \rr,  
\end{equation}
where $s_\rr$ is a real-valued \emph{scale} scalar  and $\hat{\rr}$ is an integer tensor. The scale captures the wide common range of the numerical values, and $\hat{\rr}$ encodes the actual value. 
In \modelm, $\hat{\rr}$ consists of 8-bit signed integers in $\{-128, \dots, +127\}$ or $\{0,\dots,+255\}$, depending on whether the tensor is non-negative. The Conv, ReLU, and addition operations are defined with this hybrid format, and they can be implemented efficiently with integer-only arithmetic. With a little abuse of notation, we still call numbers in this hybrid format 8-bit integers, though it can represent non-integer values with the scale scalar. 
We defer the discussion of the implementation of quantizer to Sec.~\ref{sec:optim_quant}.

\paragraph{Integer-Only Convolution}
The integer-only convolution is defined as $\yy = \conv(\xx;\WW,\bb)$, where $\WW$ is a $C\times D\times k\times k$ convolution kernel tensor with $C$, $D$ denoting the number of output / input channels, $\bb$ is a $C$-dimensional bias vector, $\xx$ is a $D\times h \times w$ input tensor, and $\yy$ is a $C\times h'\times w'$ output tensor. $(\yy, \xx, \WW)$  are all integer tensors with a floating-point scale scalar, while $\bb$ is kept in the floating-point format. 
The convolution is performed in the form
\begin{equation}
    \label{eq:conv}
    y_c = \sum_{c'=1}^D W_{c,c'}\circledast x_{c'}+b_c, \quad c\in \{1,\dots, C\}
\end{equation}
Here, $W_{c,c'}$ is a $k\times k$ $2$-D kernel, $b_c$ is a scalar, $x_{c'}$ is a $h\times w$ $2$-D input feature map, $y_c$ is a $h'\times w'$ $2$-D output feature map, and $\circledast$ denotes for 2D-convolution. In our architecture, we fix $C=D, h'=h, w'=w$ within a residual block.

Using the hybrid format defined as Eqn.~(\ref{eq:q}), we have $\yy\approx s_{\yy}\hat \yy$, $\xx\approx s_{\xx}\hat \xx$, $\WW\approx s_{\WW}\hat \WW$. Plugging them into Eqn.~(\ref{eq:conv}) yields 
%With kernel weights and inputs both represented in integers, convolution in Eqn.~\ref{eq:conv} can be written as
\begin{align*}
    s_{\yy}\hat y_c \approx y_c & = \sum_{c'=1}^D W_{c,c'}\circledast x_{c'} + b_c \approx \sum_{c'=1}^D \tilde W_{c,c'}\circledast\tilde x_{c'} + b_c \\
    & = \sum_{c'=1}^D \left(s_\WW \hat W_{c,c'}\right) \circledast \left(s_\xx \hat x_{c'}\right) + b_c \\
    & = s_\WW s_\xx \sum_{c'=1}^D \hat W_{c,c'}\circledast\hat x_{c'} + b_c.
\end{align*}

\begin{figure}[t]
\vskip 0.2in
\subfigure[Integer-arithmetic inference.]{
\begin{minipage}{0.95\columnwidth}
\centering
\includegraphics[width=\columnwidth]{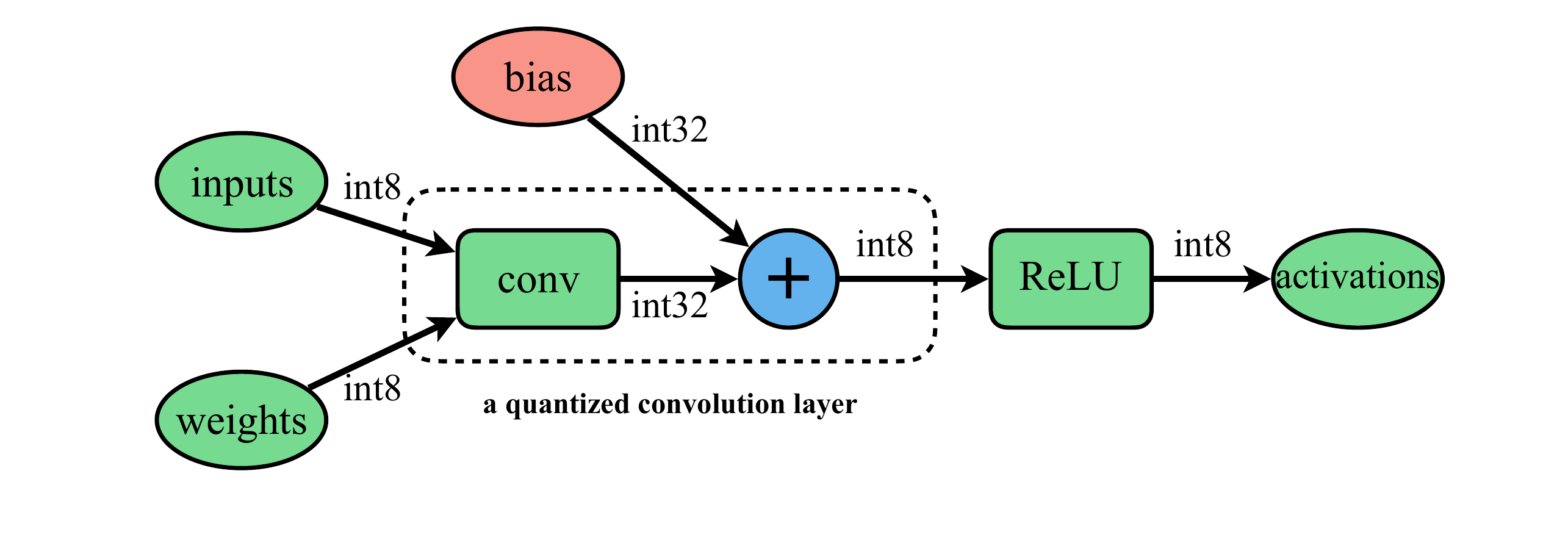}
\end{minipage}
\label{fig:true_q}
}
\subfigure[Fake quantization in training.]{
\begin{minipage}{0.95\columnwidth}
\centering
\includegraphics[width=\columnwidth]{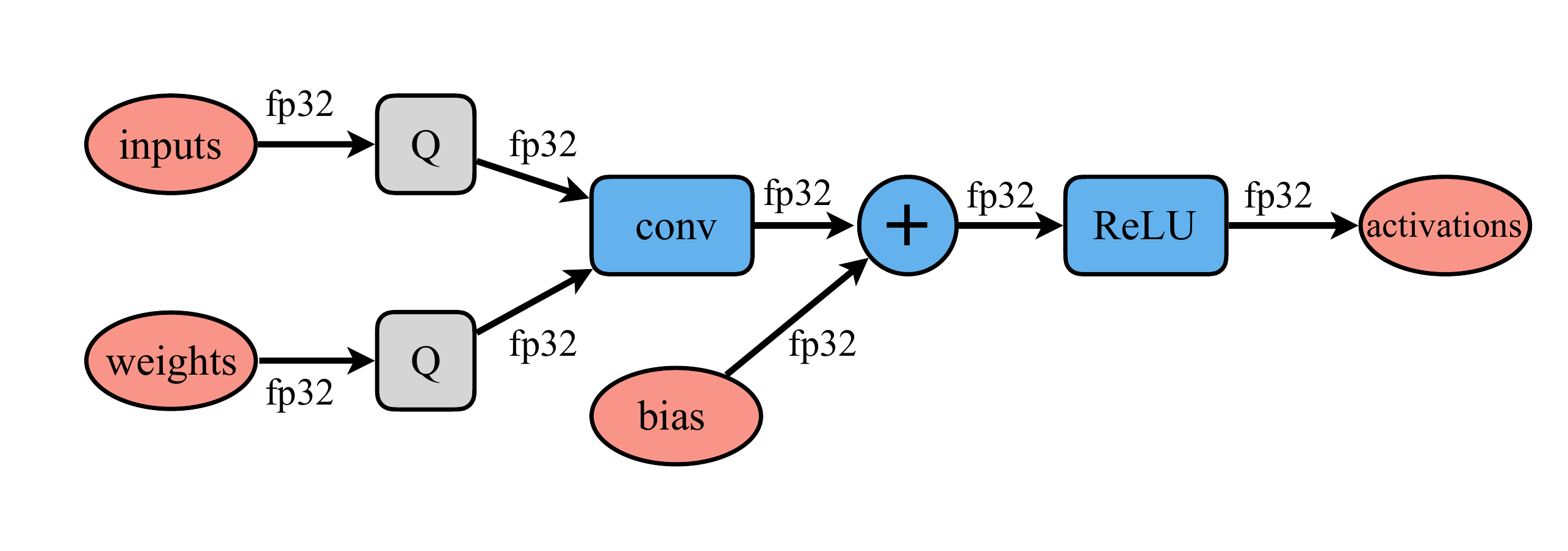}
\end{minipage}
\label{fig:fake_q}
}
\vskip -0.1in
\label{fig:quant}
\caption{True quantization and fake quantization. Best viewed in color. } 
\end{figure}

Reorganizing the terms, we have 
\begin{equation}
    \label{eq:quant_conv}
    \hat y_c \approx \frac{s_\WW s_\xx}{s_\yy} \sum_{c'=1}^D \hat W_{c,c'}\circledast\hat x_{c'} + \frac{b_c}{s_\yy}.
\end{equation}
Eqn.~(\ref{eq:quant_conv}) can be implemented as the convolution of two signed 8-bit integer tensors $\hat \WW$ and $\hat \xx$, followed by element-wise floating-point multiplication and additions. This can be realized efficiently on GPUs as a single operation in the TensorRT library, as illustrated in Figure.~\ref{fig:true_q}.  Matrix multiplications in the convolution are performed with the INT8 kernel, and the summation is performed by a 32-bit integer accumulator. Bias addition is also performed with 32-bit integers. The outputs are again quantized to an 8-bit integer using Eqn.~\ref{eq:quant_conv}. Because computations of a convolution layer are dominated by the multiplications in the summation, substituting floating-point operations with integer-arithmetic leads to a significant computation reduction when deploying on hardware. 

\paragraph{Integer-Only ReLU}
Given an input tensor $\xx \approx s_\xx \hat \xx$, the ReLU is directly applied to the signed 8-bit integer part as $\hat \yy = \mbox{ReLU}(\hat \xx) = \max\{0,\hat\xx\}$. The scale scalar is not affected $s_\yy=s_\xx$. Hence, we have $\yy \approx s_\yy\hat \yy = s_\xx \max\{0, \hat \xx\} \approx \max\{0,\xx\} = \mbox{ReLU}(\xx)$.

\subsection{Training Integer-Only Residual Blocks}
\label{sec:optim_quant}
So far, we have defined the integer-only residual blocks with a general definition of the quantizer. We have not yet discussed how to train such blocks or implement the quantizer, which we shall do now. 
As mentioned in the last subsection, integer-only residual blocks rely on quantizers to convert real-valued tensors to integers. The conversion is lossy, which usually causes inaccurate outputs of the model. Additionally, the scale parameter in Eqn.~\ref{eq:q} is crucial to the performance of quantized networks. This section focuses on fine-tuning the integer-only network with \emph{simulated quantization training} (a.k.a. fake quantization)~\cite{jacob2018quantization}.

The true integer-only inference must be deployed on hardware with special tools, which is not convenient for our training. So we implement fake quantization in PyTorch, which still uses floating-point operations but simulates the integer arithmetic by the quantizers as shown in Figure.~\ref{fig:fake_q}. This corresponds to using the convolution defined in Eqn.~(\ref{eq:conv}).

Given a known scale $s$, we define the quantizer in Eqn.~(\ref{eq:q}) in the specific form:
%than quantizer defined by Eqn.\ref{eq:q}, to keep consistent bit-width with real integer-arithmetic inference, we add a \emph{clip} operation in the quantizer, and estimate integer tensor $\hat{\rr}$ by a rounding operator
\begin{equation}
\label{eq:q2}
\begin{aligned}
    \tilde{\rr} &  = Q(\rr) = s\hat{\rr} = s\cdot\round{\mrm{clip}\left(\frac{\rr}{s}, -Q_N, Q_P\right)}.
\end{aligned}
\end{equation}
The clipping operation acts element-wise on the tensor $\rr/s$:
\begin{equation}
    \mrm{clip}\left(\frac \rr s, -Q_N, Q_P\right) = \max\left(\min\left(\frac \rr s, Q_P\right), Q_N\right).
\end{equation}
We set $Q_N=-128, Q_P=127$ for weight tensors and $Q_N=0, Q_P=255$ for non-negative activation tensors.

To optimize the network parameters with the quantizer, we leverage learned step-size quantization (LSQ) \cite{esser2019learned}. LSQ treats the scalar $s$ as a learnable parameter, which is updated by a gradient-based optimization algorithm. Back-propagation through quantizer is performed with the Straight Through Estimator (STE)\cite{bengio2013estimating} in the form $ \partial \round{\uu} / \partial \uu = \II$ for real-valued vectors $\uu$. Thus we have 
\begin{equation}
\label{eq:grad_flow}
    \frac{\partial \mathcal{L}}{\partial \rr} = \frac{\partial \mathcal{L}}{\partial \tilde{\rr}}\frac{\partial \tilde \rr }{\partial \rr} = \frac{\partial \mathcal{L}}{\partial \tilde \rr},~ \rr = \WW~or~\xx,
    % \partial \mathcal{L} / \partial \rr = (\partial\mathcal{L} / \partial \tilde \rr)(\partial \tilde \rr / \partial \rr) = \partial \mathcal{L} / \partial \tilde \rr,
\end{equation}
% Thus we have $\partial \tilde \xx / \partial \xx = \II$, $\partial \tilde \WW / \partial \WW = \II$. Here $\II$ is an identity tensor with the same shape as $\xx$ or $\WW$, respectively.\jianfei{not clear} 
where $\mathcal{L}$ denotes for the objective function and $\rr$ applies to $\WW$ or $\xx$. With Eqn.~(\ref{eq:grad_flow}) we can perform gradients back-propagation in \model normally.

For scale parameters, the gradient of $s$ can be calculated as follows,
\begin{equation*}
    \frac{\partial \tilde{\rr}}{\partial s} = \left \{ \begin{array}{ll}
          \left ( -\rr / s + \round{ \rr / s } \right ) \odot \mathbb{I}(-Q_N < \rr / s < Q_P) \\ [0.2cm]
          -Q_N \cdot \mathbb{I}(\rr / s < -Q_N) \\ [0.2cm]
          Q_P \cdot \mathbb{I}(\rr / s > Q_P)
    \end{array} \right .
\end{equation*}
where $\mathbb{I}(\cdot)$ is an indicator function  that returns a tensor of the same shape as $\rr$, and $\odot$ is the element-wise multiplication.
Applying the chain rule, we have 
\begin{equation*}
    \frac{\partial \mathcal{L}}{\partial s} = \frac{\partial \mathcal{L}}{\partial \tilde \rr}  \frac{\partial \tilde \rr}{\partial s}
\end{equation*}
We adopt the gradient re-scaling trick introduced in \cite{esser2019learned}, multiplying the gradient of $s$ by a scale factor $g=1/\sqrt{C Q_P}$, where $C$ is the number of channels. This gradient re-scaling trick helps to stabilize the learning of the scale parameter $s$.

\begin{table}[t]
\caption{Latency of floating-point and integer-only inference for  convolutions with varying number of input and output  channels. Obtained by averaging over 1000 runs (milliseconds).}
\label{tab:conv_speedup}
\vskip 0.15in
\begin{center}
\begin{small}
\begin{sc}
\begin{tabular}{cc|ccc}
\toprule
In Chn & Out Chn & Fp32 & Int8 & Speedup \\
\midrule
128 & 128 & 0.040 & 0.0039 & 10.2$\times$\\
64 & 256 & 0.035 & 0.0098 & 3.6$\times$\\
32 & 512 & 0.037 & 0.0131 & 2.8$\times$\\
\bottomrule
\end{tabular}
\end{sc}
\end{small}
\end{center}
\vskip -0.1in
\end{table}

\subsection{A More Efficient Architecture for Quantization}
\label{sec:arch}

To make the inference of the integer-only model more efficient on hardware, we improve the network architecture. We first replace dense blocks in IDF with residual blocks for their more regular architecture and fewer connections across layers. DenseNets are more memory-intensive yet less computation-intensive~\cite{zhang2021resnet} since it has many concatenation operations, which lead to many expensive quantization and dequantization operations during inference. Furthermore, DenseNets have many convolutions with a small number of input / output channels, which have unsatisfactory INT8 speedup.  Tab.~\ref{tab:conv_speedup} shows the inference latency of floating-point and INT8 convolution layers for a varying number of input / output channels. All convolutions use $3\times 3$ kernels and $16\times 16$ input / output feature maps, so their FLOPs are kept the same. The floating-point inference latency of these three convolutions is similar, but the integer inference latency differs significantly. When the channels of input and output feature maps are identical, integer arithmetic can bring better speedup than when there is a big difference between input and output channels. Convolutions in ResNets are mainly of the former shapes, while those in DenseNets are the opposite. Therefore, residual blocks are better building blocks than dense blocks for \modelm.

\subsection{Learnable Binary-Gated Convolution}
\label{sec:pruning}

\begin{figure}[t]
\vskip 0.2in
\begin{center}
\centerline{\includegraphics[width=\columnwidth]{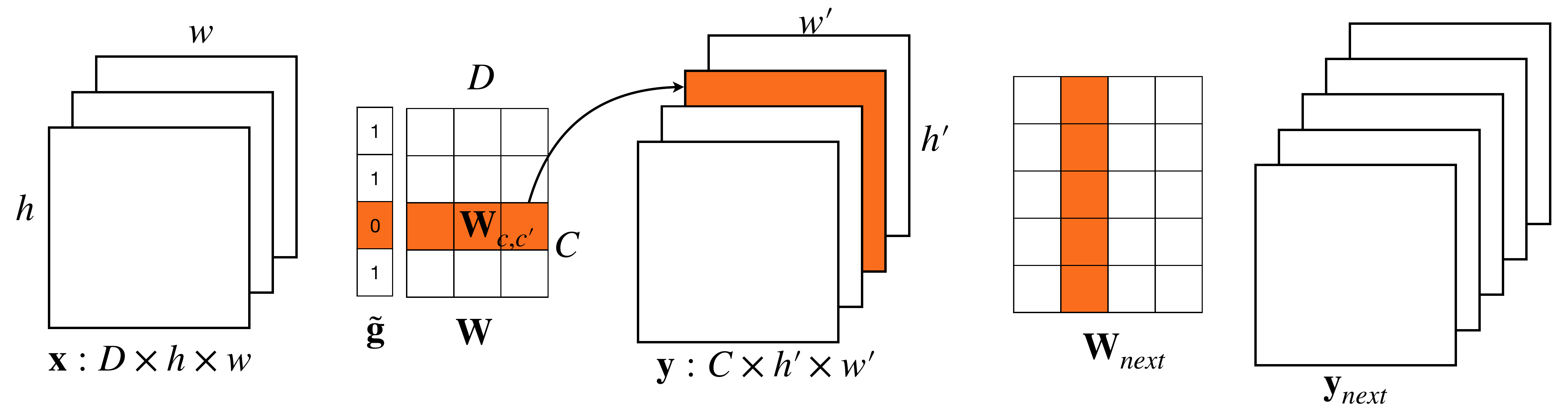}}
\caption{A gated convolution layer with $C\times D \times k \times k$ kernel weights $\WW$. The weights are represented by a $C$-by-$D$ chessboard with each element as a $2$-D $k$-by-$k$ kernel. Each gate entry $\tilde \vg$ determines whether to disable a filter (a row of kernels) in the kernel weights. Disabling a filter leads to removing the output feature map and corresponding kernels of the successive convolution that act on this feature map.}
\label{fig:prune_filters} 
\end{center}
\vskip -0.2in
\end{figure}

Neural networks have many redundant computations. Many channels die during training, and they are rarely used for inference. This problem is particularly severe for flow-based models since each transformation step has a separate network, and the required network width may vary for each transformation step. \model addresses this problem by adding learnable binary gates to integer-only convolutions, where the masked gates can be removed at the inference time.

Formally, we denote a learnable binary gate by $\tilde{\vg} = b(\vg) \coloneqq \mathbb{I}(\vg > 0.5)$, where $\vg \in [0,1]^C$.
Then a gated convolution is defined as (omitting the bias):
\begin{equation}
\begin{aligned}
     \yy & = \mrm{GConv}(\xx;\WW, \vg)\coloneqq B(\tilde \vg) \odot \mrm{Conv}(\xx;\WW)
     \\
    & 
    = \sum_{c'=1}^D \left(\tilde g_{c}  W_{c,c'}\right) \circledast x_{c'},
\end{aligned}
\end{equation}
where $B(\tilde \vg)$ is a broadcast operation to a $C\times h' \times w'$ tensor with entries $B(\tilde \vg)_{c,i,j} = \tilde g_c, \forall i=1,\dots,h', j=1,\dots,w'$. 

Figure~\ref{fig:prune_filters} illustrates the process of pruning out a filter with a binary gate. $\tilde g_c=0$ relates to disabling the filter $W_{c,:}$ and zeroing a output feature map $y_c$. Then the corresponding entries in the next convolution layer's weight that apply on this feature map are also removed. Therefore, pruning out $m$ out of $C$ filters of a convolution will reduce $m/C$ computations for both current and the next layer. 

\paragraph{Training Binary-Gated Convolution} In the direction of obtaining good gates which contain as many zeros as possible while doing little harm to the performance of the whole model, we optimize $\vg$ based on both the original training objective and the $l_1$-norm of $\tilde{\vg}$, i.e., the number of ones in the vector $\tilde{\vg}$.   Let $\mathcal{L}_{IDF}$ be the original IDF objective function, the objective of \modelm~is formalized as follows,
\begin{equation}
\label{eq:obj_gated}
\begin{aligned}
    \mathcal{L}(\XX; & \{ \WW \}, \{ \vg \}) = \mathcal{L}_{IDF} (\XX; \{\WW\}, \{\vg\}) 
     + \sum \lambda \norm{\tilde \vg}_1, 
\end{aligned}
\end{equation}
Here $\{\WW\}, \{\vg\}$ denote for the sets of all convolution kernel matrices and all gates vectors respectively.  All gate vectors $\vg$ are initialized as $\alpha \bm{1}$ with $0.5 \leq \alpha < 1$ so original gated convolution retains all filters. The $\mathcal{L}_{IDF}$ term tends to keep all entries of $\tilde \vg$ close to $1$ while $\norm{\tilde \vg}_1$ term pushes the gates to be sparse. $\lambda$ acts as a strength hyper-parameter balancing them. See Appendix~\ref{app:optim} for settings of $\lambda$ in different parts of our model. Taking derivative of $\mathcal{L}$ w.r.t. $\vg$, we have 
\begin{equation*}
    \label{ea:grad_gates}
    \frac{\partial \mathcal{L}}{\partial \vg } = \frac{\partial \tilde{\vg}}{\partial \vg} \left ( \frac{\partial \mathcal{L}_{IDF}}{\partial \tilde \vg}  + \lambda \frac{\partial \norm{\tilde \vg}_1}{\partial \tilde \vg} \right ) 
\end{equation*}
${\partial \mathcal{L}_{IDF}} / {\partial \tilde \vg}$ can be obtained by backward-propagating through the neural network. $\partial {\norm{\tilde\vg}_1} / {\partial \tilde \vg}= \bm{1}$ since $\tilde \vg$ is binary. We adopt STE to make gradients flow through the binarize operation by taking $\partial \tilde \vg / \partial \vg = \II$. Based on the gradients $\partial \mathcal{L} / \partial \vg$, gates can be optimized with gradient-based algorithms to achieve a good balance between efficiency and density modeling performance.

\paragraph{Binary-Gated Residual Blocks}
\begin{algorithm}[t]
   \caption{Training \modelm}
   \label{alg:binary_gate}
\begin{algorithmic}
   \STATE {\bfseries Input:} $r_{target}$, remaining target proportion  of FLOPs and $\XX$, the training dataset.
   \STATE \#Stage 1:
   \STATE $\WW \longleftarrow$ InitializeParameter()
   \STATE Train $\mathcal{L}_{IDF} (\XX; \{\WW\}, \{\mathbf{1}\})$ to convergence
   \STATE $F_0 \longleftarrow$ CalculateFLOPs($\WW$)
   \STATE \#Stage 2:
   \STATE $\vg \longleftarrow \alpha \bm{1},\ \lambda \longleftarrow$ InitializeLambda()
   \STATE Train $\mathcal{L} (\XX; \{\WW\}, \{\mathbf{\vg}\}) $ until CalculateFLOPs($\WW, \vg$)$<r_{target}F_0$
   \STATE \#Stage 3:
   \STATE Fine-tune $\mathcal{L}_{IDF} (\XX; \{\WW\}, \vg)$ with fixed $\vg$   
   \STATE \#Stage 4:
   \STATE Fine-tune the model with fake quantization applied to activations 
   \STATE \#Stage 5:
   \STATE Fine-tune the model with fake quantization applied to activations and weights
\end{algorithmic}
\end{algorithm}
To prune out filters, we  need to carefully consider related layers which can be influenced by the filter removal. For simple network architectures, pruning across consecutive layers is relatively straightforward, as shown in  Figure.~\ref{fig:prune_filters}. However, pruning residual blocks is a little more complicated since the addition operation in Eqn.~\ref{eq:resb} requires the two operands, i.e., the shortcut and the  convolution output, to have the same number of feature maps. This is not guaranteed in general since the convolution has a trainable gate. 
We solve this problem by performing the addition in the original unpruned feature map with a \emph{scatter add ($\mrm{SAdd}$)} operation:
\begin{align*}
    t_\theta^{(l)}(\xx) = \relu(\mrm{SAdd}(&Q(\xx), \\
    &\mrm{GConv}(\relu(\mrm{GConv}(Q(\xx))).
\end{align*}
The scatter add operation directly sums up the unpruned $Q(\xx)$ and the sparse pruned output of the gated convolution by maintaining indices representing which feature maps are removed.
As a result, convolution layers within the residual blocks can be pruned arbitrarily without considering the alignment with the shortcut path.

\subsection{Training Workflow}
We propose a 5-stage training algorithm for \modelm, as shown in Alg.~\ref{alg:binary_gate}. We first train a full-precision, non-gated model to convergence by optimizing $\mathcal{L}_{IDF} (\XX; \{\WW\}, \{\mathbf{1}\}) $. Next, we insert learnable binary gates into convolutions, set the strength hyper-parameter $\lambda$, and train the gated model by optimizing  $\mathcal{L} (\XX; \{\WW\}, \{\mathbf{\vg}\})$. After removing all zeroed-out input and output filters in $\WW$, we obtain a \emph{pruned} model. We train the gated model until the FLOPs of the pruned model reaches our target. Then, we fine-tune the pruned model by optimizing $\mathcal{L}_{IDF} (\XX; \{\WW\}, \{\mathbf{\vg}\}) $, keeping the gates fixed. The final two stages fine-tune the model to use integer-only arithmetic by incorporating fake quantization. 

\subsection{Deployment on Hardware}
So far, \model is still at the simulation level with quantizers mimicking the behaviors of integer arithmetic. To earn true inference speedup effects, we must deploy \model on specific hardware that support accelerated integer operations. As an example, NVIDIA T4 GPUs attain 16 times as many integer operations per second as floating point numbers\footnote{\url{https://www.nvidia.com/en-us/data-center/tesla-t4/}}.
In this work, the T4 GPU is selected due to more convenient software support, specifically, the TensorRT library \cite{tensorrt}. Notably, we do not rely upon any unique features of T4, so \model can be successfully deployed to other hardware and achieve the corresponding acceleration effect.

\input{5-experiments}

\section{Conclusion}
We propose Integer-Only Discrete Flows (\modelm), an efficient flow-based neural compressor. We propose a hardware-friendly backbone architecture with integer-only residual blocks. By equipping integer-arithmetic convolutions with learnable binary gates, we prune out redundant filters, significantly reducing the number of parameters and the amount of computation. Furthermore, we directly deploy \model on a Tesla T4 GPU and measure the encoding latency, showing that \model achieves up to $10\times$ speedup compared to IDF.  

\section*{Acknowledgements}

This work was supported by National Key Research and Development Project of China (No. 2021ZD0110502); NSF of China Projects (Nos. 62061136001, 61620106010, 62076145, U19B2034, U1811461, U19A2081, 6197222, 62106120); Beijing NSF Project (No. JQ19016); Beijing Outstanding Young Scientist Program NO. BJJWZYJH012019100020098; a grant from Tsinghua Institute for Guo Qiang; the NVIDIA NVAIL Program with GPU/DGX Acceleration; the High Performance Computing Center, Tsinghua University; and Major Innovation \& Planning Interdisciplinary Platform for the ``Double-First Class" Initiative, Renmin University of China.

% \newpage

\bibliography{example_paper}
\bibliographystyle{icml2022}

%%%%%%%%%%%%%%%%%%%%%%%%%%%%%%%%%%%%%%%%%%%%%%%%%%%%%%%%%%%%%%%%%%%%%%%%%%%%%%%
%%%%%%%%%%%%%%%%%%%%%%%%%%%%%%%%%%%%%%%%%%%%%%%%%%%%%%%%%%%%%%%%%%%%%%%%%%%%%%%
% APPENDIX
%%%%%%%%%%%%%%%%%%%%%%%%%%%%%%%%%%%%%%%%%%%%%%%%%%%%%%%%%%%%%%%%%%%%%%%%%%%%%%%
%%%%%%%%%%%%%%%%%%%%%%%%%%%%%%%%%%%%%%%%%%%%%%%%%%%%%%%%%%%%%%%%%%%%%%%%%%%%%%%
\newpage
\appendix
\input{appendix}

\end{document}

%% file: 5-experiments.tex
\section{Experiments}

To illustrate the efficiency and capacity of \modelm, we conduct two sets of experiments regarding to the architecture design, filter pruning, and integer-only inference. 
The models are trained with PyTorch~\cite{paszke2019pytorch} implementation and latency results are measured by deploying on a Tesla T4 GPU with the TensorRT library. 
Density estimation performance is reported in bits per dimension (bpd). 
We compare \modelm~with IDF on ImageNet32 and ImageNet64~\cite{deng2009imagenet} dataset\footnote{There are two different versions of ImageNet32 and ImageNet64 datasets.
We use the down-sampled ImageNet datasets from \url{https://image-net.org/data/downsample/Imagenet32_train.zip}, following~\citet{grcic2021densely, hazami2022efficient}. \citet{hoogeboom2019integer} use the datasets downloaded from \url{http://image-net.org/small/train_32x32.tar}. }. The flow architecture is taken from IDF, which has 3 levels of flow steps at the resolution $16\times 16$, $8\times 8$, and $4\times 4$ on ImageNet32, and 4 levels of flow steps at the resolution $32\times 32$, $16\times 16$, $8\times 8$, and $4\times 4$. Each resolution level has 8 additive coupling layers. Batch normalization is not used in both models. Following IDF, we adopt the rezero trick \cite{kingma2018glow} to realize identity mapping initialization which is helpful to improving training stability. 
The models are trained 100 epochs for ImageNet32 and 50 epochs for ImageNet64. See Appendix~\ref{app:arch} for architecture and training details. Open-source code is available at \url{https://github.com/thu-ml/IODF}.

\begin{table}[t]
\caption{Overall evaluation results on test datasets (measured in bits per dimension) of IDF-DenseNets, IDF-ResNets, and pruned models of different FLOPs pruning ratio on ImageNet32 and ImageNet64. FLOPs are measured in floating-point operations.}
\label{tab:bpd_pruned}
\vskip 0.15in
\begin{center}
\begin{small}
\begin{sc}
\begin{tabular}{lccc}
\toprule
Model & \#FLOPs & \#Parameters & bpd  \\
\midrule
ImageNet32  \\
\midrule
IDF-Dense & 6.43G & 58.4M & 3.890 \\
IDF-Res  &  7.15G & 62.2M & 3.916 \\
IDF-Res-Pruned1  & 5.67G & 38.2M &  3.916 \\
IDF-Res-Pruned2  & 4.25G & 23.5M & 3.920  \\
IDF-Res-Pruned3  & 3.28G & 17.0M &  3.930 \\
IDF-Res-Pruned4  & 1.60G & 7.5M & 4.048 \\
\midrule 
ImageNet64 \\
\midrule
IDF-Dense   & 26.27G & 84.3M & 3.629 \\
IDF-Res     & 29.09G & 84.5M & 3.630\\
IDF-Res-Pruned1 & 23.18G & 39.1M & 3.642  \\
IDF-Res-Pruned2 & 17.27G & 26.3M & 3.665 \\
IDF-Res-Pruned3 & 12.35G & 18.6M & 3.700 \\
\bottomrule
\end{tabular}
\end{sc}
\end{small}
\end{center}
\vskip -0.2in
\end{table}

\subsection{Network Architecture and Filter Pruning}
\label{exp:1}
\paragraph{Network Architecture}
IDF and IODF differ by the network architecture adopted in the additive coupling layer. 
We first compare the DenseNet architecture used in IDF (denoted as IDF-Dense) and the more hardware-friendly ResNet architecture discussed in Sec.~\ref{sec:arch} (denoted as IDF-Res). 
The models are used in full precision. 
Table~\ref{tab:bpd_pruned} compares these two architectures. IDF-Res and IDF-Dense achieve similar bpd under comparable FLOPs and number of parameters, indicating that replacing DenseNet with ResNet will not sacrifice much modeling capacity. However, IDF-Res is much more efficient than IDF-Dense with integer arithmetic, as we shall see in Sec.~\ref{exp:3}.

\paragraph{Filter Pruning} 
Next, we study the effectiveness of the learned binary gates proposed in Sec.~\ref{sec:pruning} on pruning redundant filters. 
We insert learnable binary gates into a well-trained IDF-Res model and perform training stage 2 and stage 3 depicted in Alg.~\ref{alg:binary_gate} until it satisfies targeted FLOPs and prune out $0$-gated filters as illustrated in Sec.~\ref{sec:pruning}. See Appendix \ref{app:optim} for more details about optimization parameters. We set different targeted FLOPs reduction (20\%, 40\%, 60\%, 80\% of original IDF-Res) and obtain several pruned models IDF-Res-Pruned\{1,2,3,4\}.  
Table~\ref{tab:bpd_pruned} compares the FLOPs and density estimation performance of the pruned models. Binary gated convolution can effectively reduce the number of parameters and FLOPs of IDF-Res with little harm to modeling capacity. 60\% computations can be cut with only a 0.015 bpd drop on ImageNet32. For ImageNet64, pruning is relatively harder, $57.6\%$ reduced FLOPs can cause a $0.07$ bpd increase from $3.630$ of IDF-Res to $3.700$ of IDF-Res-Pruned3.

\input{table_whole}

\begin{figure}[t]
% \vskip 0.2in
\begin{center}
\centerline{\includegraphics[width=0.9\columnwidth]{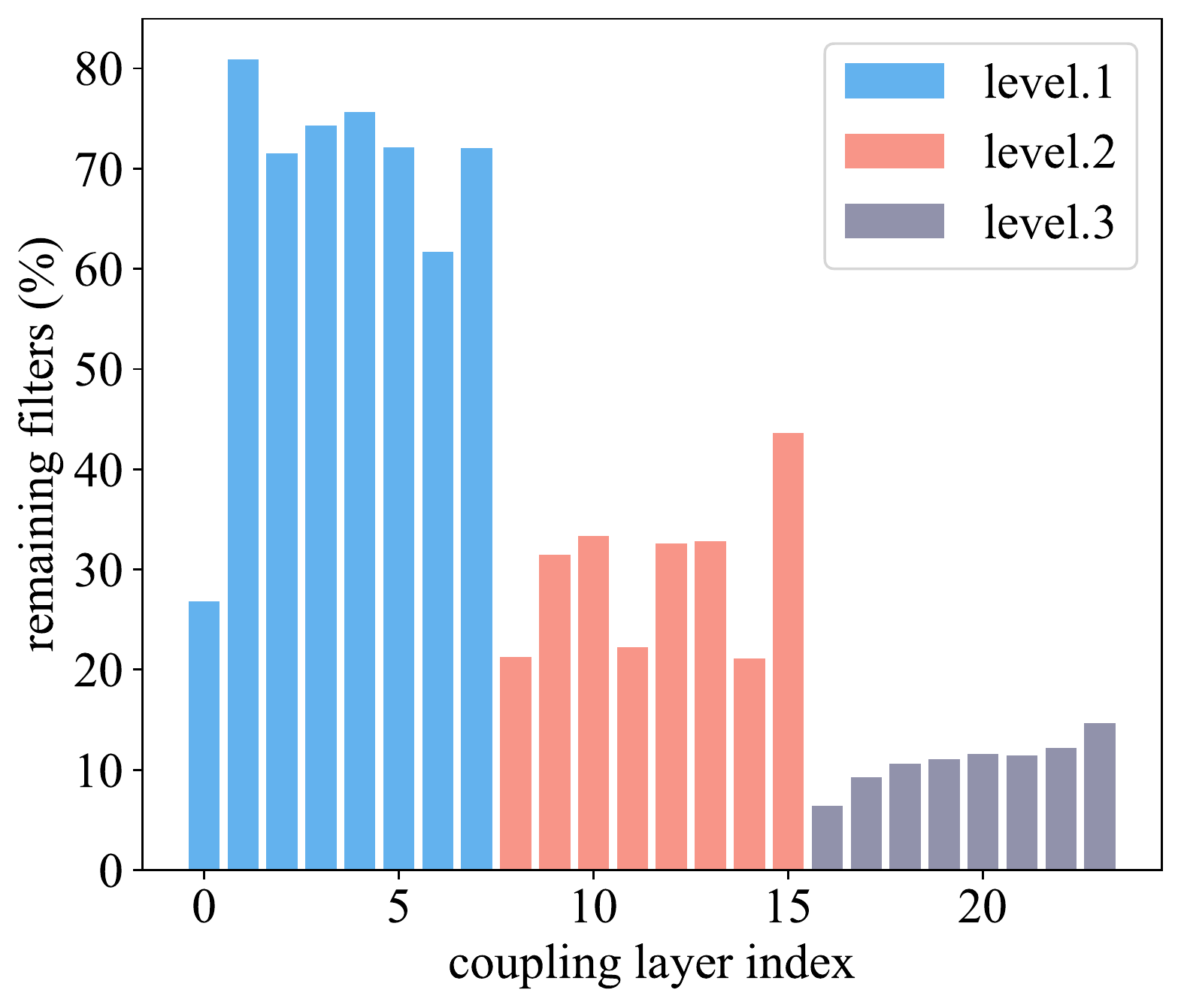}}
\caption{Number of remaining filters in different coupling layers after pruning. Layers in different flow levels are distinguished by their colors.}
\label{fig:exp_pruning}
\end{center}
\vskip -0.5in
\end{figure}

Figure~\ref{fig:exp_pruning} displays the distribution of the percentage of remained filters among different coupling layers in a pruned model trained on ImageNet32. Most filters that are pruned out concentrate in the third flow level of IDF-Res, even we set the regularization strength $\lambda$  according to the size of feature maps, i.e., $\lambda$ is larger for convolutions in the first flow level whose feature map size is $16\times 16$ and smaller for those in the third flow level whose feature map size is $4\times 4$. We observe this phenomenon in all the pruned models. 
One possible explanation is that these filters act on feature maps of very small size, which have minor functionalities
in density estimation. Another possible answer is that, due to the multi-scale architecture of the IDF model, part of the objective only relies on the first and second flow levels. Hence, parameters in these levels play more critical roles in density estimation.

\subsection{Latency Evaluation of IODF}
\label{exp:3}

Now we consider the full IODF with integer-only arithmetic. We conduct a set of experiments to display density estimation performance and inference speedup of quantized models. We use unsigned per-tensor quantization for activations and signed per-channel quantization for weight tensors\footnote{Per-tensor quantization uses one scale and offset parameters for the activation tensor. Per-channel quantization has a different scale and offset for each channel of weight tensor.}. %We insert quantizers into well-trained floating-point IDF-ResNets, IDF-DenseNets, and pruned IDF-ResNets fine-tune it with simulated quantization training as stated in Sec.~\ref{sec:optim_quant}. 
IODF is trained with the complete 5-stage procedure depicted in Alg.~\ref{alg:binary_gate}. 
The first convolution layer within each coupling
transformation step is not quantized. Afterwards, we deploy these low-precision models on a Tesla T4 GPU and evaluate their inference latency. See Appendix for the detailed environment setup. For rigorous comparison, INT8 models and FP32 models are built into inference engines with the TensorRT library. We consider the following models: (1) pure FP32 IDF-Dense and IDF-Res; (2) INT8 quantized  IDF-Dense and IDF-Res; (3) FP32 IDF-Res with half of the FLOPs pruned; and (4) \modelm, which is quantized INT8 IDF-Res with half of the FLOPs pruned. We evaluate latency for different batch sizes. 

Table~\ref{tab:latency} shows the overall results. We see that the INT8 inference of \model is $5.9\times$ faster on ImageNet32 and $8.7\times$ faster on ImageNet64 than the baseline IDF-Dense on average. \model achieves up to $10.4\times$ speedup with a batch size 16 on  ImageNet64. Comparing IDF-Res and IDF-Dense and their INT8 versions, we see that the model architecture improvement in Sec.~\ref{sec:arch} is necessary for efficient inference. Pure FP32 inference of IDF-Res is faster than IDF-Dense even the former has more parameters and FLOPs. Additionally, INT8 inference of IDF-Res is on average $5.3\times$ faster than FP32 inference while only $2.3\times$ faster for IDF-Dense. 

For larger batch sizes, inference latency per sample is lower, and the speedup effect of \model is more remarkable. This is promising in commercial applications since real-world images are mainly high resolution. Limited by hardware, training generative models is impossible on such large images directly. Thus we train the model on smaller patches and perform encoding in a patch-based manner, which naturally gives rise to a large batch size scenario.

\begin{figure}[t]
% \vskip 0.in
\begin{center}
\centerline{\includegraphics[width=\columnwidth]{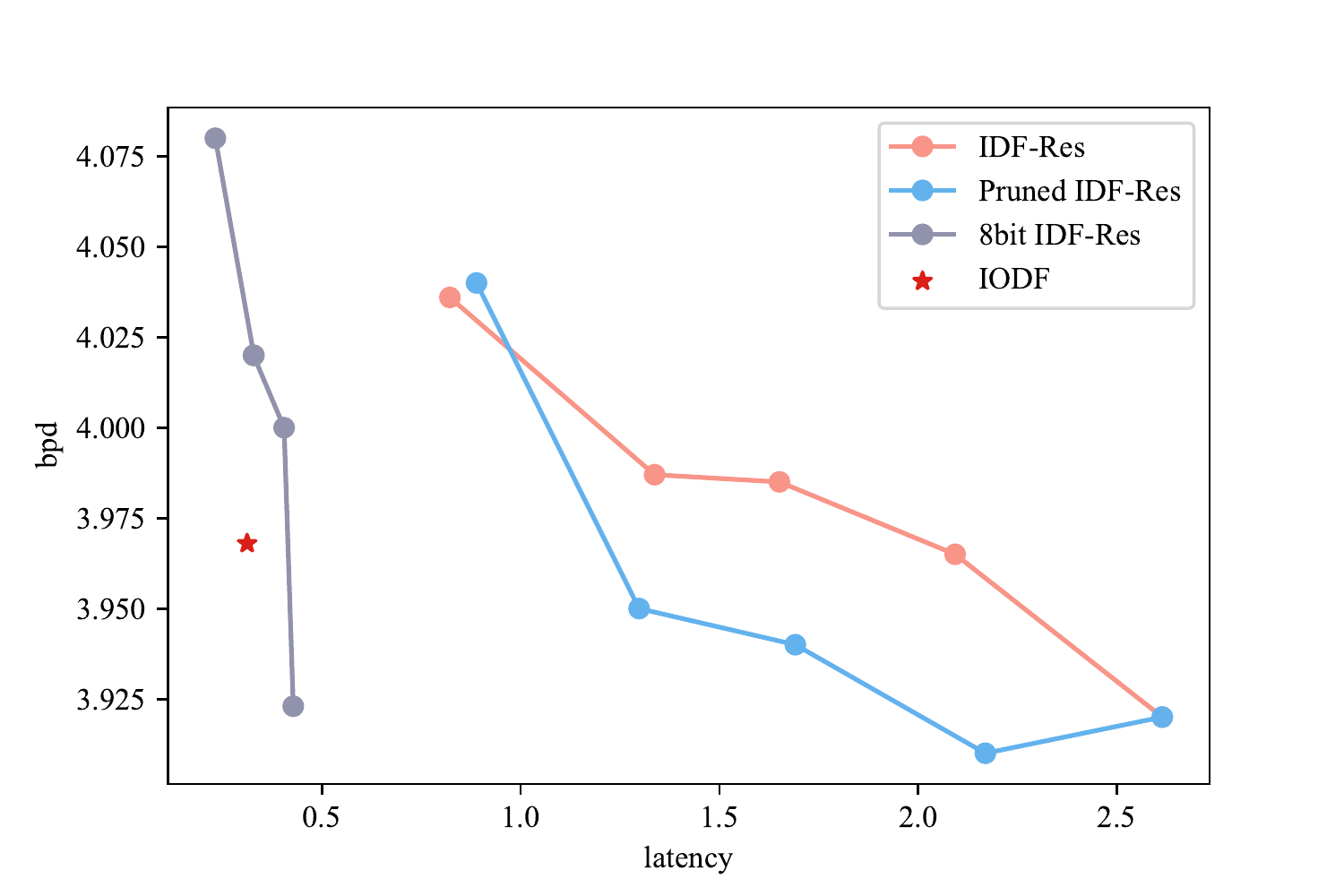}}
\caption{Bpd-latency trade-off of IDF models with a varying number of filters (IDF-Res) and their corresponding quantized models (8bit IDF-Res), different-sized models pruned from a single large IDF model (Pruned IDF-Res), and \modelm. Models are trained on ImageNet32 and evaluated with batch size=32.}
\label{fig:bpd-latency}
\end{center}
\vskip -0.3in
\end{figure}

We also implement rANS\cite{duda2009asymmetric, duda2013asymmetric} on the CPU to do actual encoding and report the actual size of compressed images as coding bpd. 
The coding BPD aligns well with the analytic bpd. 
Furthermore, the compression bandwidth is determined by both the inference latency and the cost of running rANS. %\footnote{The compression bandwidth reported here is lower than that reported in the introduction due to the difference of GPU and images to compress.} 
IODF achieves 5.6$\times$ higher bandwidth than IDF-Dense. The speedup is lower than that for inference latency due to our suboptimal CPU implementation of rANS. An optimized GPU implementation of rANS should fill the gap, which we leave as future work.

Additionally, we train IDF models with a varying number of filters (IDF-Res), prune a single large model to different extents (Pruned IDF-Res), and quantize the unpruned, different-sized IDF model (8bit IDF-Res). Figure~\ref{fig:bpd-latency} presents a bpd-latency trade-off curve of these models and shows that pruned IDF-Res and 8bit IDF-Res achieve better Pareto frontier than IDF-Res, and \model is better than both.

We also conduct experiments to evaluate memory usage, generalization ability, and practical applicability of \modelm. We evaluate the models' compression performance on the high-resolution image dataset CLIC \footnote{\url{http://compression.cc/tasks/\#image}} ($\sim 10^6$ pixels per image) \cite{agustsson2017ntire} and compare the performance with non-neural compression algorithms. As shown in Table~\ref{tab:high_reso}, the models can generalize to realistic images well. Compared to IDF, \model improves the bandwidth by ten times and reduces memory usage by 47\%, with little compression rate drop.
\model can attain better compression ratio over the traditional PNG codec~\cite{boutell1997png}, while being about 3 times slower than the CPU-based PNG compressor implemented in the \texttt{Pillow-SIMD} package. 
We also compare the speed with GPU implementations of JPEG2000 codecs,  where the bandwidth of CUJ2K and Fastvideo JPEG2000 encoders are 60.5MB/s and 985MB/s, respectively\footnote{\url{https://www.fastcompression.com/benchmarks/benchmarks-j2k.htm}}.

\begin{table}
% \vskip 0.4in
\caption{Compression performance on high resolution image dataset CLIC. (Models are trained on ImageNet64 and evaluated with batch size=32). Compression rate is measured in bpd, bandwidth in MB/s, and GPU memory usage in GB.}
\label{tab:high_reso}
\begin{center}
\begin{small}
\resizebox{\linewidth}{!}{
\begin{tabular}{l|ccccc|c}
\toprule
 Model & \makecell[c]{IDF-\\Dense} & \makecell[c]{IDF-\\Res} & \makecell[c]{8bit\\IDF-Res} & \makecell[c]{Pruned\\IDF-Res} & IODF & PNG   \\
 \midrule 
BPD & \textbf{2.438} & 2.430 & 2.499 & 2.451 &  2.505 &  3.62 \\
% \midrule
Bandwidth & 0.84 & 1.28 & 7.57 & 1.95 & \textbf{9.17} & {29.8}   \\
% Bandwidth & 0.84 & 1.28 & 7.57 & 1.95 & \textbf{9.17} & *   \\
% \makecell[l]{Memory\\Usage} \\
Memory & 3.2 & 2.8 & 1.7 & 2.4 & \textbf{1.7} & * \\
\bottomrule
\end{tabular}
}
\vskip -0.4in
\end{small}
\end{center}
\end{table}

%% file: table_whole.tex
\begin{table*}[ht]
\caption{Evaluate density  estimation, compression rate, and compression latency of IDF-DenseNets, IDF-ResNets, and \model models respectively. Likelihood and compression rate are measured in bits per dimension (raw data is 8 bits/dimension). Inference latency is measured in milliseconds per sample (lower is better). Compression bandwidth is measured in MB/s (higher is better). The last row in each part shows the speedup of \model compared to IDF-DenseNets. * denotes a failure in deployment with TensorRT.}
\label{tab:latency}
\vskip 0.15in
\begin{center}
\begin{small}
\begin{sc}
\resizebox{\linewidth}{!}{
\begin{tabular}{l|c|c|ccccc|ccccc}
\toprule
 & \multirow{2}* {\makecell[c]{Analytic\\bpd}} & \multirow{2}* {\makecell[c]{Coding\\bpd}} & \multicolumn{5}{c|}{Inference Latency} & \multicolumn{5}{c}{Compression Bandwidth} \\
 & & &  4 & 8 & 16 & 32 & 64 & 4 & 8 & 16 & 32 & 64 \\
\midrule
\multicolumn{13}{l} {ImageNet32} \\
\midrule
% \emph{PNG}  & - & 6.59  & & & & & & & & & & \\
% \emph{JPEG2000}  & - & 6.21 & & & & & & & & & & \\
% \emph{WebP} & - & 4.95 & & & & & & & & & & \\
IDF-Dense & 3.890 & 3.900 & 8.38 & 5.08 & 4.08 & * & * & 0.31 & 0.54 & 0.70 & * & *  \\
IDF-Res   & 3.916 & 3.926 & 4.19 & 3.19 & 3.59 & 2.54 & 2.93 & 0.56 & 0.79 & 0.79 & 1.12 & 1.00  \\
8bit IDF-Dense  & 3.911  & 3.921 & 5.38 & 2.90 & 1.74 & 1.20 & 0.99 & 0.46 & 0.86 & 1.21 & 2.18 & 2.67  \\
8bit IDF-Res & 3.923 & 3.934 & 2.08 & 1.09 & 0.64 & 0.44 & 0.36 & 0.91 & 1.78 & 2.96 & 4.76 & 5.98 \\
Pruned IDF-Res & 3.936 & 3.947 & 3.27 & 2.04 & 1.60 & 1.33 & 1.29 & 0.72 & 1.14 & 1.59 & 2.01 & 2.15 \\
\model & 3.968 & 3.979 & 1.79 & 0.94 & 0.54 & 0.34 & 0.27 & 0.98 & 1.91 & 3.45 & 5.41 & 7.08  \\
Speedup& - & - & 4.7\tim & 5.4\tim & 7.6\tim & * & * & 3.2\tim & 3.6\tim & 4.9\tim & * & * \\
% \model & 3.968 & 3.979 & \makecell[c]{1.79\\4.7$\times$} & \makecell[c]{0.94\\5.4$\times$} & \makecell[c]{0.54\\7.6$\times$} & 0.34 & 0.27 \\ 
\midrule 
\multicolumn{13}{l} {ImageNet64}  \\
\midrule
% \emph{PNG}  & - & 5.89 & & & & & & & & & &  \\
% \emph{JPEG2000}  & - & 4.79 &   & & & & & & & & & \\
% \emph{WebP} & - & 4.32 & & & & & & & & & & \\
IDF-Dense & 3.635 & 3.638 & 18.65 & 15.45 & 13.93 & * & * & 0.59 & 0.73 & 0.83 & * & *  \\
IDF-Res   & 3.637 & 3.640 & 12.50 & 11.89 & 9.30 & 8.84 & 8.64 & 0.82 & 0.93 & 1.22 & 1.32 & 1.35 \\
8bit IDF-Dense  & 3.663 & 3.666  & 8.98 & 5.57 & 4.35 & 3.67 & 3.34 & 1.02 & 1.66 & 2.32 & 2.83 & 3.11 \\
8bit IDF-Res & 3.663 & 3.673 & 3.03 & 2.02 & 1.61 & 1.41 & 1.31 & 1.83 & 3.47 & 4.72 & 5.57 & 6.83\\
Pruned IDF-Res & 3.657 & 3.666 & 7.75 & 6.45 & 6.55 & 5.79 & 5.71 & 1.21 & 1.59 & 1.70& 1.93 & 2.00\\
\model & 3.685 & 3.695 & 2.79 & 1.71 & 1.34 & 1.15 & 1.06 & 2.22 & 3.72 & 4.64 & 7.03 & 7.93 \\
Speedup & - & - & 6.9\tim & 9.0\tim & 10.4\tim & * & * & 3.8\tim & 5.1\tim & 5.6\tim & * & * \\
% \model & 3.685 & 3.695 &  \makecell[c]{2.79\\6.9$\times$} & \makecell[c]{1.71\\9.0$\times$} & \makecell[c]{1.34\\10.4$\times$} & 1.15 & 1.06 \\ 
\bottomrule
\end{tabular}
}
\end{sc}
\end{small}
\end{center}
\vskip -0.1in
\end{table*}

%% file: appendix.tex
\onecolumn

\section{Asymmetric Numeral System}
Asymmetric Numeral Systems (ANS) \cite{duda2009asymmetric, duda2013asymmetric} is an approach to encoding a string of discrete symbols with a known distribution into a bit stream and decoding symbols from the bit stream. ANS is a kind of arithmetic coding algorithms, achieving approximate optimal code length, i.e. entropy of the distribution. Range-base ANS (rANS) is a variant of ANS with fast coding speed. 

Let $S=(s_1, \dots, s_n)$ be the input string of symbols with each symbol taken from the alphabet set $\mathcal{A} = \{ a_1, \dots, a_k \}$. Assume the distribution over alphabet is given by $\mbf p = \{p_1, \dots, p_k\}$ with $\sum_{i=1}^k p_i = 1$. Then a large integer $M$ a chosen as total mass and integers $\{ F_{a_1}, \dots, F_{a_k}\}$ represent mass of each symbol in the alphabet, with $p_i\approx F_{a_i} / M$. Then define a cumulative mass $C_{a_i} = \sum_{j=1}^{i-1} F_{a_j}$. rANS keeps track of input symbols with a single integer state. Let $X_t$ represent the state after rANS encodes $t$ symbols in string $S$. $X_0$ is initialized to $0$. When $X_{t}$ comes, rANS update the state $X_t$ based on $X_{t-1}$ and $s_t$ in the form 
\begin{equation}
    X_{t} = \lfloor {\frac{X_{t-1}}{F_{s_{t}}}} \rfloor * M + C_{s_{t}} + X_{t-1}\smod F_{s_{t}}. 
\end{equation}
rANS decoder takes in a state $X_{t}$ and retrieves previous state $X_{t-1}$ and encoded symbol $s_t$. Consider that  
\begin{equation}
    X_{t}\smod M = C_{s_{t}} + X_{t-1}\smod F_{s_{t}}
\end{equation}
must lies in $[C_{s_t}, C_{s_t} + F_{s_t} )$. Thus the symbol $s_{t+1}$ can be retrieved by 
\begin{equation}
    s_t = a_{l} \quad C_{a_l} \leq X_t\smod M < C_{a_{l+1}}.
\end{equation}
Then 
\begin{equation}
    X_{t-1} = \lfloor \frac{X_{t-1}}{F_{s_{t}}} * F_{s_t} \rfloor + {X_{t-1}}\smod F_{s_t} = \lfloor \frac{X_t}{M} \rfloor * F_{s_t}+ (X_t\smod M - C_{s_t}) 
\end{equation}

\section{Experimental Details}

\subsection{Network Architecture}
\label{app:arch}
\begin{figure}[h]
\label{fig:resnets}
\vskip 0.2in
\begin{center}
\centerline{\includegraphics[width=\columnwidth]{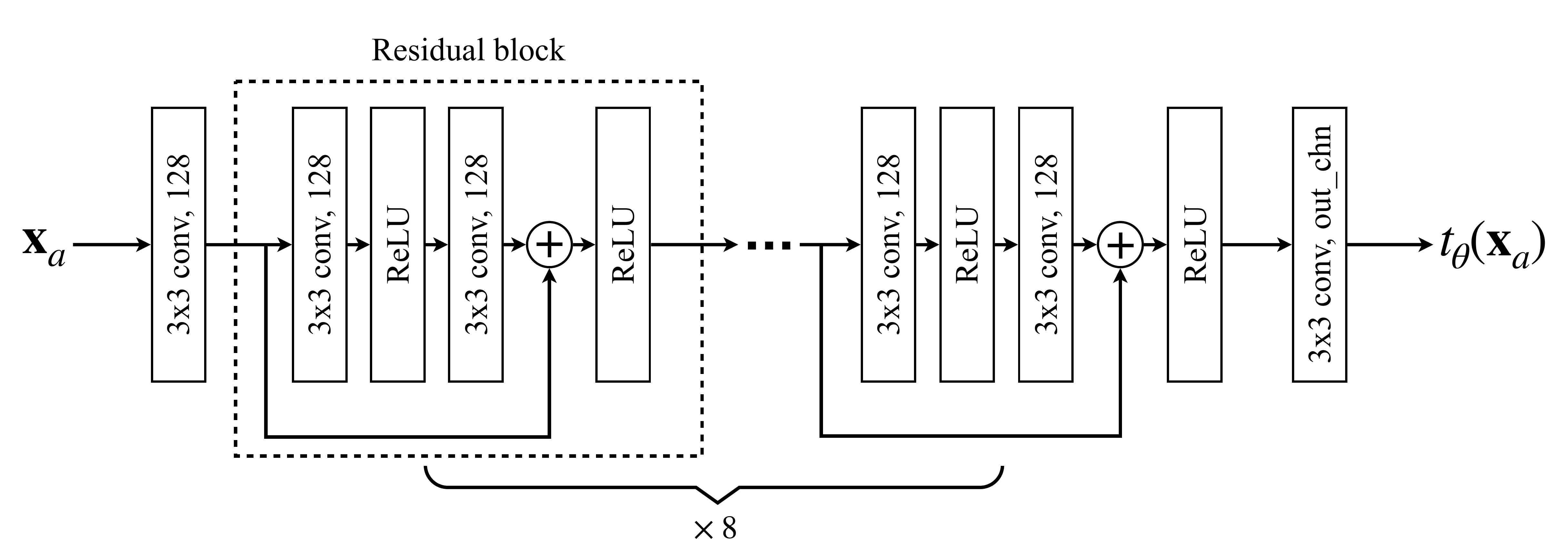}}
\caption{$t_\theta(\cdot)$ in each coupling transformation consist of $8$ residual blocks and two convolutions. All convolutions use $3\times 3$ kernel and 128 hidden channels. We set \emph{padding=1} and \emph{stride=1} in convolution for not changing the shape of feature maps.}
\label{fig:network_arch}
\end{center}
\vskip -0.2in
\end{figure}

The overall architecture of \model is the same as IDF introduced in Sec.~\ref{sec:idf}. The entire invertible transformation from $x$ to $z$ has $L$ levels and each level is composed of $D$ coupling layers. The neural network $t_\theta(\cdot)$ in each coupling layer consists of $8$ residual blocks as Figure.~\ref{fig:network_arch} shows. 

The architecture for IDF-ResNets and optimization parameters are shown in Tab.~\ref{tab:exp}.
\begin{table}[t]
\caption{IDF-ResNets architecture and optimization parameters for each experiment.}
\label{tab:exp}
\vskip 0.15in
\begin{center}
\begin{small}
\begin{sc}
\begin{tabular}{lcccccccr}
\toprule
Dataset & L & D  & batchsize & train samples & optimizer & lr &  lr decay & epochs \\
\midrule 
ImageNet32 & 3 & 128 & 512 & 1230000 & Adamax & 0.001 & 0.99 & 100 \\ 
ImageNet64 & 4 & 128 & 256 & 1230000 & Adamax & 0.0001  & 0.99 & 50 \\
\bottomrule
\end{tabular}
\end{sc}
\end{small}
\end{center}
\vskip -0.2in
\end{table}

\subsection{Optimization Parameters for Binary Gates and LSQ}
\label{app:optim}
In training model with gated convolutions, we initialize gates with $\alpha=0.8$ and set $lr=0.00005,\ lr\_decay=0.99$. We set strength parameter $\lambda$ in Eqn.~\ref{eq:obj_gated} according to which layer the convolution locates in, as shown in Tab.~\ref{tab:lambda}. In fine-tuning pruned model, we set $lr = 0.00005,\ lr\_decay=0.99$. Model with gated convolutions is trained for 50 epochs and pruned models are fine-tuned for 5 epochs.

\begin{table}[t]
\caption{We choose larger strength parameters for gated convolutions in shallower level.}
\label{tab:lambda}
\vskip 0.15in
\begin{center}
\begin{small}
\begin{sc}
\begin{tabular}{lcccc}
\toprule
Dataset & level 1 & level 2 & level 3 & level 4 \\
\midrule 
ImageNet32 & 1 & 2 & 4 & - \\ 
ImageNet64 & 1 & 2 & 4 & 8 \\
\bottomrule
\end{tabular}
\end{sc}
\end{small}
\end{center}
\vskip -0.2in
\end{table}

For quantized model, we initialize scale parameters in quantizers data dependently in the form 
\begin{equation}
    s_\rr = \frac{2\frac{1}{n_{\rr}}\sum |r|}{\sqrt {2^b-1}},~~\rr=\WW~or~\xx, 
\end{equation}
where $b$ is the bit width and $n_\xx$ denotes for the number of elements in tensor $\xx$. We set $lr=1e-4,\ lr\_decay=0.99$ in simulated quantization training and quantized models are fine-tuned for 10 epochs. 

\subsection{Hardware and Software}
\label{app:hardware}
The codes for our experiments are implemented with PyTorch \cite{paszke2019pytorch}. The model implementation is based on IDF codes released by \cite{hoogeboom2019integer}. rANS implementation is based on local bits back code released by \cite{ho2019compression} in C language.  

We train \model using 8 Nvidia RTX 2080Ti GPUs. We build inference engine and evaluate the latency on a Tesla T4 GPU and Intel(R) Xeon(R) Platinum 8259CL CPU @ 2.50GHz with TensorRT8.2.0.6, CUDA10.2.

%% file: example_paper.bbl
\begin{thebibliography}{58}
\providecommand{\natexlab}[1]{#1}
\providecommand{\url}[1]{\texttt{#1}}
\expandafter\ifx\csname urlstyle\endcsname\relax
  \providecommand{\doi}[1]{doi: #1}\else
  \providecommand{\doi}{doi: \begingroup \urlstyle{rm}\Url}\fi

\bibitem[Agustsson \& Timofte(2017)Agustsson and Timofte]{agustsson2017ntire}
Agustsson, E. and Timofte, R.
\newblock Ntire 2017 challenge on single image super-resolution: Dataset and
  study.
\newblock In \emph{\cvpr~Workshops}, pp.\  126--135, 2017.

\bibitem[Ball{\'e} et~al.(2018)Ball{\'e}, Johnston, and
  Minnen]{balle2018integer}
Ball{\'e}, J., Johnston, N., and Minnen, D.
\newblock Integer networks for data compression with latent-variable models.
\newblock In \emph{\iclr}, 2018.

\bibitem[Bengio et~al.(2013)Bengio, L{\'e}onard, and
  Courville]{bengio2013estimating}
Bengio, Y., L{\'e}onard, N., and Courville, A.
\newblock Estimating or propagating gradients through stochastic neurons for
  conditional computation.
\newblock \emph{arXiv preprint arXiv:1308.3432}, 2013.

\bibitem[Bird et~al.(2021)Bird, Kingma, and Barber]{bird2020reducing}
Bird, T., Kingma, F.~H., and Barber, D.
\newblock Reducing the computational cost of deep generative models with binary
  neural networks.
\newblock In \emph{\iclr}, 2021.

\bibitem[Boutell \& Lane(1997)Boutell and Lane]{boutell1997png}
Boutell, T. and Lane, T.
\newblock Png (portable network graphics) specification version 1.0.
\newblock \emph{Network Working Group}, pp.\  1--102, 1997.

\bibitem[Chen et~al.(2020{\natexlab{a}})Chen, Gai, Yao, Mahoney, and
  Gonzalez]{chen2020statistical}
Chen, J., Gai, Y., Yao, Z., Mahoney, M.~W., and Gonzalez, J.~E.
\newblock A statistical framework for low-bitwidth training of deep neural
  networks.
\newblock In \emph{\nips}, 2020{\natexlab{a}}.

\bibitem[Chen et~al.(2020{\natexlab{b}})Chen, Lu, Chenli, Zhu, and
  Tian]{chen2020vflow}
Chen, J., Lu, C., Chenli, B., Zhu, J., and Tian, T.
\newblock Vflow: More expressive generative flows with variational data
  augmentation.
\newblock In \emph{\icml}, pp.\  1660--1669. PMLR, 2020{\natexlab{b}}.

\bibitem[Choi et~al.(2018)Choi, Wang, Venkataramani, Chuang, Srinivasan, and
  Gopalakrishnan]{choi2018pact}
Choi, J., Wang, Z., Venkataramani, S., Chuang, P. I.-J., Srinivasan, V., and
  Gopalakrishnan, K.
\newblock Pact: Parameterized clipping activation for quantized neural
  networks.
\newblock \emph{arXiv preprint arXiv:1805.06085}, 2018.

\bibitem[Courbariaux et~al.(2015)Courbariaux, Bengio, and
  David]{courbariaux2015binaryconnect}
Courbariaux, M., Bengio, Y., and David, J.-P.
\newblock Binaryconnect: Training deep neural networks with binary weights
  during propagations.
\newblock In \emph{\nips}, pp.\  3123--3131, 2015.

\bibitem[Deng et~al.(2009)Deng, Dong, Socher, Li, Li, and
  Fei-Fei]{deng2009imagenet}
Deng, J., Dong, W., Socher, R., Li, L.-J., Li, K., and Fei-Fei, L.
\newblock Imagenet: A large-scale hierarchical image database.
\newblock In \emph{\cvpr}, pp.\  248--255. IEEE, 2009.

\bibitem[Dinh et~al.(2017)Dinh, Sohl-Dickstein, and Bengio]{dinh2016density}
Dinh, L., Sohl-Dickstein, J., and Bengio, S.
\newblock Density estimation using real nvp.
\newblock In \emph{\iclr}, 2017.

\bibitem[Domke et~al.(2008)Domke, Karapurkar, and Aloimonos]{4587817}
Domke, J., Karapurkar, A., and Aloimonos, Y.
\newblock Who killed the directed model?
\newblock In \emph{\cvpr}, pp.\  1--8, 2008.

\bibitem[Dong et~al.(2019)Dong, Yao, Gholami, Mahoney, and
  Keutzer]{dong2019hawq}
Dong, Z., Yao, Z., Gholami, A., Mahoney, M.~W., and Keutzer, K.
\newblock Hawq: Hessian aware quantization of neural networks with
  mixed-precision.
\newblock In \emph{\cvpr}, pp.\  293--302, 2019.

\bibitem[Duda(2009)]{duda2009asymmetric}
Duda, J.
\newblock Asymmetric numeral systems.
\newblock \emph{arXiv preprint arXiv:0902.0271}, 2009.

\bibitem[Duda(2013)]{duda2013asymmetric}
Duda, J.
\newblock Asymmetric numeral systems: entropy coding combining speed of
  {H}uffman coding with compression rate of arithmetic coding.
\newblock \emph{arXiv preprint arXiv:1311.2540}, 2013.

\bibitem[Esser et~al.(2020)Esser, McKinstry, Bablani, Appuswamy, and
  Modha]{esser2019learned}
Esser, S.~K., McKinstry, J.~L., Bablani, D., Appuswamy, R., and Modha, D.~S.
\newblock Learned step size quantization.
\newblock In \emph{\iclr}, 2020.

\bibitem[Frankle \& Carbin(2019)Frankle and Carbin]{frankle2018lottery}
Frankle, J. and Carbin, M.
\newblock The lottery ticket hypothesis: Finding sparse, trainable neural
  networks.
\newblock In \emph{\iclr}, 2019.

\bibitem[Grci{\'c} et~al.(2021)Grci{\'c}, Grubi{\v{s}}i{\'c}, and
  {\v{S}}egvi{\'c}]{grcic2021densely}
Grci{\'c}, M., Grubi{\v{s}}i{\'c}, I., and {\v{S}}egvi{\'c}, S.
\newblock Densely connected normalizing flows.
\newblock In \emph{\nips}, volume~34, pp.\  23968--23982, 2021.

\bibitem[Group et~al.(2000)]{joint2000jpeg2000}
Group, J. P.~E. et~al.
\newblock Jpeg2000 image coding system.
\newblock \emph{ISO/IEC FCD 15444-1}, 2000.

\bibitem[Han et~al.(2015)Han, Pool, Tran, and Dally]{han2015learning}
Han, S., Pool, J., Tran, J., and Dally, W.~J.
\newblock Learning both weights and connections for efficient neural networks.
\newblock In \emph{\nips}, 2015.

\bibitem[Hazami et~al.(2022)Hazami, Mama, and
  Thurairatnam]{hazami2022efficient}
Hazami, L., Mama, R., and Thurairatnam, R.
\newblock Efficient-vdvae: Less is more.
\newblock \emph{arXiv preprint arXiv:2203.13751}, 2022.

\bibitem[He et~al.(2016)He, Zhang, Ren, and Sun]{he2016deep}
He, K., Zhang, X., Ren, S., and Sun, J.
\newblock Deep residual learning for image recognition.
\newblock In \emph{\cvpr}, pp.\  770--778, 2016.

\bibitem[He et~al.(2017)He, Zhang, and Sun]{he2017channel}
He, Y., Zhang, X., and Sun, J.
\newblock Channel pruning for accelerating very deep neural networks.
\newblock In \emph{\cvpr}, pp.\  1389--1397, 2017.

\bibitem[Ho et~al.(2019{\natexlab{a}})Ho, Chen, Srinivas, Duan, and
  Abbeel]{ho2019flow++}
Ho, J., Chen, X., Srinivas, A., Duan, Y., and Abbeel, P.
\newblock Flow++: Improving flow-based generative models with variational
  dequantization and architecture design.
\newblock In \emph{\icml}, pp.\  2722--2730. PMLR, 2019{\natexlab{a}}.

\bibitem[Ho et~al.(2019{\natexlab{b}})Ho, Lohn, and Abbeel]{ho2019compression}
Ho, J., Lohn, E., and Abbeel, P.
\newblock Compression with flows via local bits-back coding.
\newblock In \emph{\nips}, 2019{\natexlab{b}}.

\bibitem[Ho et~al.(2020)Ho, Jain, and Abbeel]{ho2020denoising}
Ho, J., Jain, A., and Abbeel, P.
\newblock Denoising diffusion probabilistic models.
\newblock In \emph{\nips}, 2020.

\bibitem[Hoogeboom et~al.(2019)Hoogeboom, Peters, van~den Berg, and
  Welling]{hoogeboom2019integer}
Hoogeboom, E., Peters, J., van~den Berg, R., and Welling, M.
\newblock Integer discrete flows and lossless compression.
\newblock In \emph{\nips}, volume~32, 2019.

\bibitem[Huang et~al.(2017)Huang, Liu, Van Der~Maaten, and
  Weinberger]{huang2017densely}
Huang, G., Liu, Z., Van Der~Maaten, L., and Weinberger, K.~Q.
\newblock Densely connected convolutional networks.
\newblock In \emph{\cvpr}, pp.\  4700--4708, 2017.

\bibitem[Hubara et~al.(2016)Hubara, Courbariaux, Soudry, El-Yaniv, and
  Bengio]{courbariaux2016binarized}
Hubara, I., Courbariaux, M., Soudry, D., El-Yaniv, R., and Bengio, Y.
\newblock Binarized neural networks.
\newblock In \emph{\nips}, volume~29, 2016.

\bibitem[Huffman(1952)]{huffman1952method}
Huffman, D.~A.
\newblock A method for the construction of minimum-redundancy codes.
\newblock \emph{Proceedings of the IRE}, 40\penalty0 (9):\penalty0 1098--1101,
  1952.

\bibitem[Jacob et~al.(2018)Jacob, Kligys, Chen, Zhu, Tang, Howard, Adam, and
  Kalenichenko]{jacob2018quantization}
Jacob, B., Kligys, S., Chen, B., Zhu, M., Tang, M., Howard, A., Adam, H., and
  Kalenichenko, D.
\newblock Quantization and training of neural networks for efficient
  integer-arithmetic-only inference.
\newblock In \emph{\cvpr}, pp.\  2704--2713, 2018.

\bibitem[Kingma \& Dhariwal(2018)Kingma and Dhariwal]{kingma2018glow}
Kingma, D.~P. and Dhariwal, P.
\newblock Glow: Generative flow with invertible 1x1 convolutions.
\newblock In \emph{\nips}, 2018.

\bibitem[Kingma \& Welling(2013)Kingma and Welling]{Kingma2014AutoEncodingVB}
Kingma, D.~P. and Welling, M.
\newblock Auto-encoding variational bayes.
\newblock In \emph{\iclr}, 2013.

\bibitem[Kingma et~al.(2021)Kingma, Salimans, Poole, and
  Ho]{kingma2021variational}
Kingma, D.~P., Salimans, T., Poole, B., and Ho, J.
\newblock Variational diffusion models.
\newblock In \emph{\nips}, 2021.

\bibitem[Larochelle \& Murray(2011)Larochelle and Murray]{larochelle2011neural}
Larochelle, H. and Murray, I.
\newblock The neural autoregressive distribution estimator.
\newblock In \emph{Proceedings of the fourteenth international conference on
  artificial intelligence and statistics}, pp.\  29--37. JMLR Workshop and
  Conference Proceedings, 2011.

\bibitem[Lebedev \& Lempitsky(2016)Lebedev and Lempitsky]{lebedev2016fast}
Lebedev, V. and Lempitsky, V.
\newblock Fast convnets using group-wise brain damage.
\newblock In \emph{\cvpr}, pp.\  2554--2564, 2016.

\bibitem[LeCun et~al.(1990)LeCun, Denker, and Solla]{lecun1990optimal}
LeCun, Y., Denker, J.~S., and Solla, S.~A.
\newblock Optimal brain damage.
\newblock In \emph{\nips}, pp.\  598--605, 1990.

\bibitem[Li et~al.(2017)Li, Kadav, Durdanovic, Samet, and Graf]{li2016pruning}
Li, H., Kadav, A., Durdanovic, I., Samet, H., and Graf, H.~P.
\newblock Pruning filters for efficient convnets.
\newblock In \emph{\iclr}, 2017.

\bibitem[Lu et~al.(2021)Lu, Chen, Li, Wang, and Zhu]{lu2021implicit}
Lu, C., Chen, J., Li, C., Wang, Q., and Zhu, J.
\newblock Implicit normalizing flows.
\newblock In \emph{\iclr}, 2021.

\bibitem[Maal{\o}e et~al.(2019)Maal{\o}e, Fraccaro, Li{\'e}vin, and
  Winther]{maaloe2019biva}
Maal{\o}e, L., Fraccaro, M., Li{\'e}vin, V., and Winther, O.
\newblock Biva: A very deep hierarchy of latent variables for generative
  modeling.
\newblock In \emph{\nips}, volume~32, 2019.

\bibitem[NVIDIA(2018)]{tensorrt}
NVIDIA.
\newblock Tensorrt.
\newblock \url{https://developer.nvidia.com/tensorrt}, 2018.

\bibitem[Paszke et~al.(2019)Paszke, Gross, Massa, Lerer, Bradbury, Chanan,
  Killeen, Lin, Gimelshein, Antiga, et~al.]{paszke2019pytorch}
Paszke, A., Gross, S., Massa, F., Lerer, A., Bradbury, J., Chanan, G., Killeen,
  T., Lin, Z., Gimelshein, N., Antiga, L., et~al.
\newblock Pytorch: An imperative style, high-performance deep learning library.
\newblock In \emph{\nips}, volume~32, pp.\  8026--8037, 2019.

\bibitem[Rastegari et~al.(2016)Rastegari, Ordonez, Redmon, and
  Farhadi]{rastegari2016xnor}
Rastegari, M., Ordonez, V., Redmon, J., and Farhadi, A.
\newblock Xnor-net: Imagenet classification using binary convolutional neural
  networks.
\newblock In \emph{European Conference on Computer Vision}, pp.\  525--542.
  Springer, 2016.

\bibitem[Rezende et~al.(2014)Rezende, Mohamed, and
  Wierstra]{rezende2014stochastic}
Rezende, D.~J., Mohamed, S., and Wierstra, D.
\newblock Stochastic backpropagation and approximate inference in deep
  generative models.
\newblock In \emph{\icml}, pp.\  1278--1286. PMLR, 2014.

\bibitem[Salimans et~al.(2017)Salimans, Karpathy, Chen, and
  Kingma]{salimans2017pixelcnn++}
Salimans, T., Karpathy, A., Chen, X., and Kingma, D.~P.
\newblock Pixelcnn++: Improving the pixelcnn with discretized logistic mixture
  likelihood and other modifications.
\newblock In \emph{\iclr}, 2017.

\bibitem[Shannon(1948)]{shannon1948mathematical}
Shannon, C.~E.
\newblock A mathematical theory of communication.
\newblock \emph{The Bell system technical journal}, 27\penalty0 (3):\penalty0
  379--423, 1948.

\bibitem[Sneyers \& Wuille(2016)Sneyers and Wuille]{sneyers2016flif}
Sneyers, J. and Wuille, P.
\newblock Flif: Free lossless image format based on maniac compression.
\newblock In \emph{2016 IEEE international conference on image processing
  (ICIP)}, pp.\  66--70. IEEE, 2016.

\bibitem[Sohl-Dickstein et~al.(2015)Sohl-Dickstein, Weiss, Maheswaranathan, and
  Ganguli]{sohl2015deep}
Sohl-Dickstein, J., Weiss, E., Maheswaranathan, N., and Ganguli, S.
\newblock Deep unsupervised learning using nonequilibrium thermodynamics.
\newblock In \emph{\icml}, pp.\  2256--2265. PMLR, 2015.

\bibitem[Song et~al.(2021)Song, Meng, and Ermon]{song2020denoising}
Song, J., Meng, C., and Ermon, S.
\newblock Denoising diffusion implicit models.
\newblock In \emph{\iclr}, 2021.

\bibitem[Townsend et~al.(2019{\natexlab{a}})Townsend, Bird, and
  Barber]{townsend2019practical}
Townsend, J., Bird, T., and Barber, D.
\newblock Practical lossless compression with latent variables using bits back
  coding.
\newblock In \emph{\iclr}, 2019{\natexlab{a}}.

\bibitem[Townsend et~al.(2019{\natexlab{b}})Townsend, Bird, Kunze, and
  Barber]{townsend2019hilloc}
Townsend, J., Bird, T., Kunze, J., and Barber, D.
\newblock Hilloc: Lossless image compression with hierarchical latent variable
  models.
\newblock In \emph{\iclr}, 2019{\natexlab{b}}.

\bibitem[Van~Baalen et~al.(2020)Van~Baalen, Louizos, Nagel, Amjad, Wang,
  Blankevoort, and Welling]{van2020bayesian}
Van~Baalen, M., Louizos, C., Nagel, M., Amjad, R.~A., Wang, Y., Blankevoort,
  T., and Welling, M.
\newblock Bayesian bits: Unifying quantization and pruning.
\newblock In \emph{\nips}, volume~33, pp.\  5741--5752, 2020.

\bibitem[van~den Berg et~al.(2020)van~den Berg, Gritsenko, Dehghani,
  S{\o}nderby, and Salimans]{van2020idf++}
van~den Berg, R., Gritsenko, A.~A., Dehghani, M., S{\o}nderby, C.~K., and
  Salimans, T.
\newblock Idf++: Analyzing and improving integer discrete flows for lossless
  compression.
\newblock In \emph{\iclr}, 2020.

\bibitem[Wen et~al.(2016)Wen, Wu, Wang, Chen, and Li]{wen2016learning}
Wen, W., Wu, C., Wang, Y., Chen, Y., and Li, H.
\newblock Learning structured sparsity in deep neural networks.
\newblock \emph{\nips}, 29:\penalty0 2074--2082, 2016.

\bibitem[Zhang et~al.(2021{\natexlab{a}})Zhang, Benz, Argaw, Lee, Kim, Rameau,
  Bazin, and Kweon]{zhang2021resnet}
Zhang, C., Benz, P., Argaw, D.~M., Lee, S., Kim, J., Rameau, F., Bazin, J.-C.,
  and Kweon, I.~S.
\newblock Resnet or densenet? introducing dense shortcuts to resnet.
\newblock In \emph{\cvpr}, pp.\  3550--3559, 2021{\natexlab{a}}.

\bibitem[Zhang et~al.(2021{\natexlab{b}})Zhang, Kang, Ryder, and
  Li]{zhang2021iflow}
Zhang, S., Kang, N., Ryder, T., and Li, Z.
\newblock iflow: Numerically invertible flows for efficient lossless
  compression via a uniform coder.
\newblock In \emph{\nips}, volume~34, 2021{\natexlab{b}}.

\bibitem[Zhang et~al.(2021{\natexlab{c}})Zhang, Zhang, Kang, and
  Li]{zhang2021ivpf}
Zhang, S., Zhang, C., Kang, N., and Li, Z.
\newblock ivpf: Numerical invertible volume preserving flow for efficient
  lossless compression.
\newblock In \emph{\cvpr}, pp.\  620--629, 2021{\natexlab{c}}.

\bibitem[Zhou et~al.(2016)Zhou, Wu, Ni, Zhou, Wen, and Zou]{zhou2016dorefa}
Zhou, S., Wu, Y., Ni, Z., Zhou, X., Wen, H., and Zou, Y.
\newblock Dorefa-net: Training low bitwidth convolutional neural networks with
  low bitwidth gradients.
\newblock \emph{arXiv preprint arXiv:1606.06160}, 2016.

\end{thebibliography}
